\documentclass{aims}
\usepackage{subfigure}
\usepackage{amsmath}
\usepackage{amsfonts}
\usepackage{bbm}
\usepackage{arydshln}

  \usepackage{paralist}
  \usepackage{graphics} %% add this and next lines if pictures should be in esp format
  \usepackage{epsfig} %For pictures: screened artwork should be set up with an 85 or 100 line screen
\usepackage{graphicx}  
\usepackage{epstopdf}%This is to transfer .eps figure to .pdf figure; please compile your paper using PDFLeTex or PDFTeXify.
\usepackage[usenames,dvipsnames]{pstricks}
\usepackage{color}
 \usepackage[colorlinks=true]{hyperref}
\hypersetup{urlcolor=blue, citecolor=red}

\def\currentvolume{X}
 \def\currentissue{X}
  \def\currentyear{200X}
   \def\currentmonth{XX}
    \def\ppages{X--XX}
     \def\DOI{10.3934/xx.xx.xx.xx}

\theoremstyle{definition}

\newcommand{\classvec}[1]{\boldsymbol{\kappa}(\mathcal{H}_\text{{#1}})}
\newcommand{\classGDR}[1]{\boldsymbol{\kappa}_\text{GDR}(\mathcal{H}_\text{{#1}})}

\title[Semi-supervised classification on graphs] {
Semi-supervised classification on graphs using explicit diffusion dynamics }

\author[R.\ L.\ Peach, A.\ Arnaudon and M.\ Barahona]{}

\subjclass{Primary: ??; Secondary: ??.}
 \keywords{??}

 \email{alexis.arnaudon@imperial.ac.uk}
 \email{r.peach13@imperial.ac.uk}
 \email{m.barahona@imperial.ac.uk}

\thanks{All authors acknowledge funding through EPSRC award EP/N014529/1 supporting the EPSRC Centre for Mathematics of Precision Healthcare at Imperial.}

\thanks{$^*$Corresponding author: Mauricio Barahona}

\DeclareMathOperator*{\argmax}{argmax}

\renewcommand{\AA}[1]{\textcolor{magenta}{\textbf{[AA: #1]}}}

\begin{document}

\maketitle

% Enter the first author's name and address:
\centerline{\scshape Robert L. Peach}
\medskip
{\footnotesize
% please put the address of the first author
 \centerline{Department of Mathematics and Imperial College Business School}
   \centerline{Imperial College London}
   \centerline{London SW7 2AZ, UK}
}
\medskip
\centerline{\scshape Alexis Arnaudon and Mauricio Barahona$^*$}
\medskip
{\footnotesize
% please put the address of the first author
 \centerline{Department of Mathematics }
   \centerline{Imperial College London}
   \centerline{London SW7 2AZ, UK}
} 

\medskip

\bigskip

% The name of the associate editor will be entered by an editorial staff
% "Communicated by the associate editor name" is not needed for special issue.
% \centerline{(Communicated by the associate editor name)}

\begin{abstract}
Classification tasks based on feature vectors can be significantly improved by
including within deep learning a graph that summarises pairwise relationships between the samples.
Intuitively, the graph acts as a conduit to channel and bias the inference of class labels. 
Here, we study classification methods that consider the graph as the originator of an explicit graph diffusion. 
We show that appending graph diffusion to feature-based learning as an \textit{a posteriori} refinement achieves state-of-the-art classification accuracy. This method, which we call Graph Diffusion Reclassification (GDR), uses overshooting events of a diffusive graph dynamics to reclassify individual nodes. The method uses intrinsic measures of node influence, which are distinct for each node, and allows the evaluation of the relationship and importance of features and graph for classification.
We also present diff-GCN, a simple extension of Graph Convolutional Neural Network (GCN) architectures that leverages explicit diffusion dynamics, and allows the natural use of directed graphs. To showcase our methods, we use benchmark datasets of documents with associated citation data.
\end{abstract}

\setcounter{tocdepth}{1}
\tableofcontents

\section{Introduction}
\label{sec:introduction}

In recent years, supervised learning has become a standard tool in data science, and a variety of algorithms have been developed and applied to a range of problems~\cite{lecun2015deep,goodfellow2016deep}. One such task is classification, whereby the known labels of a training dataset are used to infer the free parameters of a model (a `classifier') in order to predict the unknown labels (or \textit{classes}) of other samples. Classic examples of such methods include the multi-layer perceptron (MLP), support vector machine (SVM) and random forest (RF), among many others~\cite{bishop2006pattern}.
In many situations there is only a small number of known labels, as compared to the number of labels to be learnt.
To improve algorithmic performance, one can leverage additional information that may be available from the unlabeled samples using, e.g., generative models~\cite{kingma2014semi}, low density separation~\cite{chapelle2005semi} or heuristic approaches~\cite{zhu2003semi}. 

In some cases, the  additional information about the dataset is in the form of pairwise relationships between samples. 
Such relationships can be formalised as a weighted graph, where the nodes are samples and the edges represent similarities. 
The ensuing graph encodes global information about the full dataset (including the unlabeled samples), and can be used to guide the training of the classifier, e.g., by including the Laplacian or adjacency matrix of the graph within the composition of features in the deep learning layers~\cite{bronstein2017geometric}. In this way, the graph can guide the learning of the inferred parameters to reflect the additional information, thus leading to improved classification. Algorithms that take advantage of an associated graph to improve supervised learning come under the banner of \textit{graph semi-supervised learning}, and include graph Laplacian regularisation and graph embedding approaches, such as label propagation (LP)~\cite{Zhu2003} and semi-supervised embedding (SemiEmb)~\cite{Weston2012}, DeepWalk~\cite{Perozzi2014} or Planetoid~\cite{Yang2016}, among others. 
Recently, state-of-the-art performance was achieved through the introduction of graph convolutional neural network (GCN) architectures based on spectral representations of graph convolutions~\cite{hammond2011wavelets,defferrard2016convolutional,Bruna2013,Kipf2016}.  Further improvements in performance have been achieved through tuning of the convolutional operators and the deep learning architectures of GCNs~\cite{Velickovic2017,Levie2018,Chen2017,Zhuang2018,Zhang2018,Gao2018,Liu2018}. 

Although graph semi-supervised learning sits at the interface of machine learning and graph theory, algorithmic developments have focussed predominantly on the deep learning architecture, with less consideration devoted to exploiting relevant graph theoretical properties, and the relationship between feature-based and graph-based learning.
For instance, for GCNs to improve classification it is necessary that the graph is \textit{aligned} with the similarities between sample feature vectors~\cite{qian2019quantifying}. 
As discussed above, using a graph within GCNs is posited as a means to bias the \textit{propagation} of label information between similar samples.  
However, standard GCN algorithms do not consider explicitly the dynamic diffusion of information on the graph, neither do they take advantge of the intrinsic inhomogeneities in the scale of sample neighbourhoods that naturally emerge from the diffusion dynamics.

Here, we present semi-supervised graph classification methods that view the graph as a generator of an explicit diffusive dynamics through its Laplacian. 
We start by exploiting this concept in its simplest form by using a continuous-time graph diffusion as a means to induce an \textit{a posteriori} re-classification of any previously obtained  assignment,
be it based on features alone, or on graph and features combined.
We take this prior probabilistic assignment as the initial condition of a diffusion dynamics, and we search the time evolution of node dynamics for large overshoots. This is the basis to relabel nodes to a different class.
Hence the probabilistic output of a classifier is filtered \textit{a posteriori} through a graph diffusion to obtain a re-labelling consistent with the graph structure. 
The unfolding provided by the graph diffusion has been used to capture features at multiple scales to detect communities in an unsupervised manner~\cite{Delvenne2010}, and to obtain graph embeddings at different scales~\cite{Schaub2019}.
More recently, overshooting dynamics has been proposed as a basis to generalise the notion of graph centrality across scales~\cite{MSC}. 
Here, we use overshooting in the graph diffusion~\cite{bacik2016flow,MSC} to capture the influence of nodes across all scales in the graph as a means to refine class labels. Note that this method affords distinct scales for each node, since overshooting events occur at different times for different nodes. We refer to this method as \textit{Graph Diffusion Re-classification} (GDR). Tests on benchmark datasets of text documents for which a citation graph exist show that GDR achieves state-of-the-art performance.
  
%Beyond the use of diffusion to induce \textit{a posteriori} re-classification through GDR, 
We also show that a graph diffusion kernel with a time hyperparameters can be included within the GCN inference.  
This simple extension of GCN, which we denote \textit{diffusive GCN} (diff-GCN), affords additional flexibility through the inferred diffusion timescale and outperforms GCN for the benchmark datasets. 
Furthermore, the use of explicit graph diffusion lends itself naturally to the case of directed graphs~\cite{bacik2016flow,beguerisse_vangelov_RMST2013}, thus allowing the use of the asymmetry of the relationships between samples for classification purposes. We exemplify the use of graph directionality on a directed citation network.

\subsection*{Structure of the paper.}

In section~\ref{sec:semi-supervised_classification} we give definitions and set up the problem; we review some classic methods of supervised and semi-supervised classification with and without a graph; and we describe the benchmark datasets used throughout.
Section~\ref{sec:GDR-section} describes the GDR algorithm that leverages explicit diffusion on the graph to reclassify nodes given a prior class assignment, and we show the result of its application to the benchmarks. 
In section \ref{sec:MT_vs_GCN}, we propose an extension to GCN where the continuous diffusion operator is embedded within the GCN inference algorithm and we illustrate its performance.  Section~\ref{sec:directed_graph} presents how the ideas behind graph diffusion extend naturally to the case when the graph is directed, and we show how this additional information can be used to improve classification. We finish in Section~\ref{sec:conclusion} with a discussion and some concluding remarks.  

\section{The use of graphs in semi-supervised classification}\label{sec:semi-supervised_classification}

\subsection*{Notation and statement of the problem}

Let us consider a set of $N$ samples $\{X_i\}_{i=1}^N$,  each described by an $F$-dimensional feature vector $X_i\in \mathbb R^F$.
Each sample belongs to one of $c$ classes and its membership is represented through an indicator vector $H_i \in [0,1]^c, \boldsymbol{1}^T H_i =1$, where $\boldsymbol{1}$ is the vector of ones.
Within the sample set, we have a \textit{training subset} of $\widetilde n< N$ samples, for which the class labels $H_i$ are known. The class labels are unknown for the remaining $n=N-\widetilde{n}$ samples. 

Given the feature vectors $X_i$ for all $N$ samples and the class labels $H_i$ for the $\widetilde{n}$ samples in the training set, the task of \textit{supervised classification} is to obtain a class assignment for the $n$ unknown labels.  

Let us assume that we have access to pairwise similarities between all samples. Then we can define a graph $G = (V,E)$ with $N=|V|$ vertices and $M = |E|$ edges, where each node corresponds to a sample $X_i$ and the edge between nodes $i$ and $j$ has a weight $A_{ij}$ reflecting the similarity between the samples $X_i$ and $X_j$.
The similarities thus form the weighted adjacency matrix $A_{N \times N}$ of the graph. We first consider the case of symmetric similarities $A=A^T$, i.e., when the graph is \textit{undirected}. The case of directed graphs is presented in Section~\ref{sec:directed_graph}.
Another important matrix associated with the graph is the Laplacian: $L_{N \times N}= D-A$, where $D=\text{diag}(A \boldsymbol{1})$ is the diagonal matrix of weighted degrees. 

The problem of \textit{semi-supervised graph classification} uses the graph $G$ together with the features $X_i$ of the full dataset and the known classes of the training set to predict the classes of the remaining nodes in the graph.

For notational compactness, we arrange our samples into two data matrices with rows given by the feature vectors $X_i$: $\widetilde{X}_{\widetilde{n} \times F}$ corresponds to the training subset and $X_{n \times F}$ corresponds to the samples with unknown class labels.
Similarly, we compile the known class labels into a $0$-$1$ membership matrix $\widetilde{H}_{\widetilde{n}\times c}$, with rows given by the membership vectors $H_i$ of the training set.  
Given $X, \widetilde{X}, \widetilde{H}$ (and potentially the graph adjacency matrix $A$), our classification task is therefore to infer a row-stochastic membership assignment matrix $H_{n \times c}$ for the samples with unknown class labels.

The assignment matrix can be hard ($0$-$1$) or probabilistic. 
A given probabilistic assignment $H$ can be turned into a hard assignment by taking the highest probability class, collected in a node assignment vector
\begin{align}
    \boldsymbol{\kappa}(H)  := \argmax_{1 \leq j \leq c} H\, , 
    \label{prior-assignment}
\end{align}
where the $\argmax$ operator is applied row-wise. 

\subsubsection*{Some additional matrix operations}

We use standard normalisations, defined here for completeness. Given a matrix $W$, we define its row-normalised version as
\begin{align}
\label{eq:1norm}
    \text{L1norm}(W) :=\text{diag}(W \boldsymbol{1})^{-1} W\, .
\end{align}
The row-wise softmax, widely used in machine learning, can then be written as
\begin{align}
\label{eq:softmax}
    \text{softmax}(W)= \text{L1norm}\left(\exp[W]\right)\, , 
\end{align}
where the matrix $\exp[W]$ represents \textit{element-wise} exponentiation, i.e., $\exp[W]_{ij}:=\exp(W_{ij})$.
By construction, both $\text{L1norm}(W)$ and $\text{softmax}(W)$ are row stochastic matrices, i.e. $\text{L1norm}(W) \boldsymbol{1} = \text{softmax}(W) \boldsymbol{1} = \boldsymbol{1}$.

Some deep learning methods mentioned below use nonlinear processing units. In particular, the rectified linear unit (ReLU) function for matrix $W$ is defined as
\begin{align}
    \text{ReLU}(W)= \frac{W + |W|}{2} = \mathrm{max}(0,W)\, ,
\end{align}
where $|W|$ and $\mathrm{max}(0,W)$ are element-wise matrix operators, i.e., $|W|_{ij} = |W_{ij}|$ and  $\mathrm{max}(0,W)_{ij}=\mathrm{max}(0,W_{ij})$.

We now provide brief descriptions of supervised and semi-supervised classification methods, without and with the use of a graph, which will serve as comparisons and prior classifiers for our diffusion-based re-classification algorithm (GDR) and the diffusive extension of GCN (diff-GCN).

\subsection{Supervised classification without a graph.} \label{supervised-classification}

\subsubsection*{Projection (no learning)}

Perhaps the simplest approach to supervised classification (without a graph) is to classify the samples with unknown labels according to their projection on the centroids of the known classes of the training set. The projection operator is easily obtained from the data and membership matrices without any parametric inference or `learning'. 

The centroids of the $c$ classes of the training data are given by the rows of the matrix
\begin{align}
\Pi_{c \times F}(\widetilde{X}) = \left(\widetilde{H}^T \widetilde{H}\right)^{-1} \widetilde{H}^T \widetilde{X} = \widetilde{H}^+ \widetilde{X}\, ,
\end{align}
where $\widetilde{H}^+$ denotes the pseudo-inverse of $\widetilde H$.
Therefore, the projection of the remaining samples onto the centroids of the classes derived from the training set is simply $X \Pi^T$, and a $n \times c$ probabilistic membership matrix for $X$ is obtained by the row normalisation
\begin{align}
\label{eq:assign_proj}
    H_\text{proj}= \text{L1norm}\left(X \Pi^T\right)\, ,
\end{align}
where the element $(H_\text{proj})_{i \mu}$ gives the probability that the sample $i$ belongs to the class $\mu$.

\subsubsection*{Supervised classifiers (with learning)}

Beyond a naive projection, many alternatives have been proposed for the problem of supervised classification through learning. These methods entail the definition of a classifier, i.e., a model with a particular structure (`architecture') and parameters to be inferred (`learnt') from the training set.
Here, we exemplify our work with three standard, widely used algorithms: the multi-layer perceptron (MLP), support vector machine (SVM), and random forest (RF) algorithms~\cite{goodfellow2016deep}. 
Similarly to~\eqref{eq:assign_proj}, each classifier outputs a $n \times c$ probabilistic assignment matrix for the unknown class labels.  For instance, an MLP classifier based on a two-layer perceptron with $d_1$ hidden units gives
\begin{align}
    \label{eq:assign_MLP}
    H_\text{MLP}= \mathrm{softmax} \left(\mathrm{ReLU}\left(X W^{0}\right)\,W^{1} \right)\, ,
\end{align}
where $W^0_{F \times d_1}(\widetilde{X})$ and $W^1_{d_1 \times c}(\widetilde{X})$ are inter-layer connectivity matrices containing the learnt parameters of the model, inferred through the optimisation of a loss function measuring the error of the prediction on the training data $\widetilde{X}$. 

The other two classifiers considered here (SVM, RF) use different heuristics to infer the corresponding assignment matrices $H_\text{SVM}$ and $H_\text{RF}$, respectively. Here, we choose these classic supervised classifiers to illustrate our work, but any other method could be used~\cite{goodfellow2016deep}.

\subsection{Semi-supervised classification with a graph}\label{GCN-section}

Given a weighted graph $G$ with adjacency matrix $A_{N \times N}$ encoding pairwise similarities between all samples, several methods have been proposed to use this additional information during the inference of the model to enhance performance~\cite{Zhu2003, Weston2012, Perozzi2014}. These methods, which are usually classed as semi-supervised graph classification algorithms, are transductive, i.e., they use information from the \textit{full} dataset during the training phase, since the graph $G$ includes relationships between all samples including those with unknown class labels.
Here we focus on the graph convolutional neural network (GCN) architecture~\cite{Kipf2016}, which has been shown to outperform other semi-supervised graph classifiers. As a comparison, we also show results from Planetoid~\cite{Yang2016}, a method that learns a graph embedding from features.

\subsubsection*{Graph convolutional neural networks (GCNs)}

GCNs originate from work in spectral graph convolutions and convolutional neural networks~\cite{defferrard2016convolutional,hammond2011wavelets}.
The GCN architecture~\cite{Kipf2016} is akin to other deep learning classifiers such as the MLP, i.e., layers of nonlinear units interconnected through matrices to be inferred through gradient learning, yet also including a graph convolution operation for every layer.

Consider an undirected graph with adjacency matrix $A$, and let us define the matrix $\mathcal{X}_{N \times F}=\begin{bmatrix} \widetilde{X} \\ X \end{bmatrix}$ containing the feature vectors of all samples.
In a GCN with two convolutional layers and $d_1$ hidden units, the $n \times c$ classifier $H_\text{GCN}$ for the samples with unknown class labels has the form~\cite{Kipf2016}:
\begin{align}
\begin{bmatrix} * 
\\  H_\text{GCN} \end{bmatrix} = \text{softmax}\left(\widehat{A} \
    \text{ReLU} \left(\widehat{A} \,
    \mathcal{X} \,
    W^0 \right) W^1 \right)\, ,
    \label{eq:fullgcn}
\end{align}
where $W^0_{F \times d_1}(\widetilde{X})$ and $W^1_{d_1 \times c}(\widetilde{X})$ are inter-layer connectivity matrices to be inferred by optimising a loss function that only includes the training nodes, $\widetilde{X}$.
The topology of the graph is included through a symmetrised transition matrix of an associated graph with self loops:
\begin{align}
\label{eq:gcn_graph_loops}
\widehat{A}&=\bar{D}^{1/2} (\bar{D}^{-1}\bar{A}) \bar{D}^{-1/2} 
= \bar{D}^{-1/2} \bar{A} \bar{D}^{-1/2} 
\end{align}
where  the adjacency matrix of the graph with self loops is $\bar{A} = I_N + A$, with $I_N$ the identity matrix of size $N$, and $\bar{D}=\text{diag}(\bar{A} \boldsymbol{1})$.
See~\cite{Kipf2016} for further details.

Note that from a dynamical perspective, the convolution in each layer of~\eqref{eq:fullgcn}--\eqref{eq:gcn_graph_loops} conveys a discrete diffusion process, since $(\bar{D}^{-1}\bar{A})$
is the transition matrix of a discrete-time Markov chain on the graph with self-loops 
Hence, each layer of the GCN applies a one-step random walk to the output of the nonlinear units. 
The training phase can thus be thought of as propagating the label information on the graph so as to bias the inference of the parameters in $W^0$ and $W^1$ through the graph structure. Below we exploit this dynamical viewpoint in more detail through an explicit formulation of the diffusive dynamics on the graph.

\section{Graph Diffusion Reclassification (GDR)}\label{sec:GDR-section}

Let $\mathcal{H}_{N \times c} = \begin{bmatrix} \widetilde{H} \\ H \end{bmatrix}$ be a row-stochastic class assignment matrix for all the samples, where $\widetilde{H}$ contains the known labels of the training set and $H$ is the probabilistic assignment for the remaining samples obtained using any of the methods described above (or any other).
The method of graph diffusion reclassification (GDR) uses $\mathcal{H}$ as the initial condition of a diffusion dynamics, and uses a particular feature of the ensuing time evolution (i.e., the presence of overshoots) to implement sample re-classification.  

\subsection{Reclassification based on overshooting of diffusive dynamics}

Let us consider a diffusive process on the graph~\cite{lambiotte2014random,masuda2017random,Schaub2019} governed by the (combinatorial) graph Laplacian
\begin{align}
    \partial_t \mathbf{p} = -L \, \mathbf{p} \, , 
    \label{network_diffusion}
\end{align}
where the vector $\mathbf{p}_{N \times 1}(t)$ is defined on the nodes of the graph. 
This linear equation is solved by the matrix exponential
\begin{align}
    \mathbf{p}(t) =e^{-t L} \, \mathbf{p}_0\, , 
    \label{diffusion-solution}
\end{align}
where $\mathbf p_0$ is the initial condition. 
If the graph is undirected, the Laplacian is a symmetric and doubly stochastic matrix, and the dynamics~\eqref{diffusion-solution} preserves the $1$-norm.
Hence the stationary state is the constant vector $\boldsymbol{\pi}= \frac{\|\mathbf{p}_0\|_1}{N} \boldsymbol{1} =: \pi \boldsymbol{1}$.

\subsubsection*{Overshooting}

The exponential operator in~\eqref{diffusion-solution} has been used as a means to reveal information about a graph across different scales~\cite{coifman2006diffusion,saerens2007,lambiotte2014random,Schaub2019}.
Here, in contrast, we use a multiscale notion also emanating from the diffusive dynamics, which captures the influence of each node on any other node of the graph~\cite{bacik2016flow,MSC}, i.e., the presence of overshooting in the approach of the dynamics~\eqref{diffusion-solution} to stationarity.

To illustrate this phenomenon, consider as initial condition a delta impulse of mass $m$ at node $i$: $\mathbf p_0 = \|\mathbf{p}_0\|_1 \, \mathbf{e}_i$, where $\mathbf{e}_i$ is the indicator vector at node $i$.
The $j$-th coordinate of the solution~\eqref{diffusion-solution} gives the time evolution at node $j$:
$p_j(t | \mathbf{e}_i )= \|\mathbf{p}_0\|_1 \left(e^{-tL}\right)_{ji}, \enskip j=1,\ldots,N$. 
For the source node, $p_i(t | \mathbf{e}_i)$, decays towards the stationary value 
$\pi$. 
For all other nodes, the value of $p_j(t| \mathbf{e}_i), \forall j \neq i$ increases from zero towards $\pi$ in two qualitative ways:
if node $j$ is closely influenced by the source, $p_j(t| \mathbf{e}_i)$ undergoes an overshoot (i.e., it crosses the stationary value); 
if node $j$ does not feel strongly the influence of the source, then $p_j(t| \mathbf{e}_i)$ approaches $\pi$ monotonically from below.  
Note that, depending on the relative connectivity graph, an overshoot can happen at different times for different nodes, i.e., it is possible to observe a `late' overshoot. The presence or absence of an overshoot (with respect to a node) over the whole time scale thus establishes a measure of influence of the node based on the diffusion on the graph. 
Given an initial class assignment $\mathbf{p}_0$ on the graph, the node overshootings are obtained from the condition
\begin{align}
    \max_{t > t_\mathrm{min}} \left (e^{-t L} \, \mathbf{p_0} - \frac{\|\mathbf{p}_0\|_1}{N} \boldsymbol{1} \right) > 0\, ,
        \label{peak-reachability}
\end{align}
where $t_\mathrm{min}$ is a burn-in time to allow the decay of the dominance of the initial class assignment.

We refer to~\cite{MSC} for a more extended discussion of this notion and its use to define a multiscale node centrality measure in graphs.  

\subsubsection*{The GDR reclassifier}

Starting with a prior assignment matrix $\mathcal{H}$, a node is reclassified according to the largest overshoot induced by any of the $c$ classes. The values of all the node overshoots are captured compactly in matrix form as
\begin{align}
\Omega(\mathcal{H}, t_\text{min}) = 
\text{ReLU}\left(
\max_{t \geq t_\text{min}} \left [\left(e^{-t L} - \frac{\boldsymbol{1} \boldsymbol{1}^T}{N} \right) \mathcal{H} \right ] \right) \, ,
\label{eq:overshoot}
\end{align}
and the reclassification of the nodes is given from the maximum overshoot across classes
\begin{align}
    \label{eq:reclassified_overshoot}
    \boldsymbol{\kappa}_\Omega(\mathcal{H}, t_\text{min}) = \argmax_{1 \leq j \leq c} \, \Omega(\mathcal{H}, t_\text{min})\, , 
\end{align}
where the $\argmax$ is a row-wise operator finding the maximum across classes, and we define $\argmax(\boldsymbol{0})=0$ so that the indicator vector $\mathbbm{1}_0(\boldsymbol{\kappa}_\Omega)$ marks the set of non-overshooting nodes.

This reclassification vector is then used to update the prior (hard) assignment $\boldsymbol{\kappa}(\mathcal{H})$ to give the GDR assignment
\begin{align}
    \label{eq:GDR_assignment}
    \boldsymbol{\kappa}_\text{GDR}(\mathcal{H}, t_\text{min})= \boldsymbol{\kappa}(\mathcal{H}) \, \text{diag}\left(\mathbbm{1}_0(\boldsymbol{\kappa}_\Omega)\right) + \boldsymbol{\kappa}_\Omega(\mathcal{H}, t_\text{min})\, .
\end{align}
Clearly, the training set is never reclassified.
The burn-in time $t_\mathrm{min}$ is a hyperparameter which is tuned for each dataset using a validation subset. 

\subsection{Application of GDR to benchmark datasets} \label{sec:results}

\subsubsection*{Datasets}

To test our models, we closely follow the experimental setups in~\cite{Yang2016,Kipf2016,qian2019quantifying}. We use three benchmark datasets~\cite{Sen2008} consisting of scientific articles Citeseer, Cora and Pubmed. Each document is described by a feature vector summarising its text, and belongs to a scientific topic (class).  For each dataset, we also have access to the associated citation network (undirected).
In addition, we use a Wikipedia dataset collected in~\cite{qian2019quantifying} consisting of $\sim20,000$ Wikipedia articles from $12$ subcategories, where the feature vectors are bag-of-words representations of the text, and the graph represents hyperlink citations. The Wikipedia dataset is an example where, contrary to the other three datasets, the classification task is not aided by the combination of graph and features~\cite{qian2019quantifying}.
Details of these datasets are summarised in Table~\ref{tbl:data}.

\begin{table}[h]
    \begin{tabular}{lccccc}
        %\hline
        \textbf{Datasets} & \textbf{Nodes} & \textbf{Edges} & \textbf{Classes} & \textbf{Features}\\
        \hline
        Citeseer & $3,327$ & $4,732$ & $6$ & $3,703$\\
        Cora & $2,708$ & $5,429$ & $7$ & $1,433$\\
        Pubmed & $19,717$ & $44,338$ & $3$ & $500$\\
        Wikipedia & $20,525$ & $215,056$ & $12$ & $100$ \\ 
       \hline
\\
    \end{tabular}
    \caption{Statistics of datasets as reported in~\cite{Yang2016}~and~\cite{qian2019quantifying}.}
    \label{tbl:data}
\end{table}

\subsubsection*{Numerical experiments}

We have used these datasets to test the performance of the reclassified vector $\classGDR{}$ as compared to the hard prior assignment $\classvec{}$ from supervised classifiers without a graph (projection, RF, SVM, MLP) and from semi-supervised graph classifiers (Planetoid and GCN) presented in Sections~\ref{supervised-classification}--\ref{GCN-section}.  In order to test the improvement due to the graph information, we have also considered a uniform prior $H_\text{unif}=(1/c)\, \boldsymbol{1} \boldsymbol{1}^T$ across samples, which ignores the information from the features of the samples. 
Each dataset was split into training, validation and test sets at different ratios, where the training set percentage of total samples were 3.6\%, 5.2\%, 0.3\% and 3.5\% for Citeseer, Cora, Pubmed and Wikipedia, respectively. 

\begin{table}[htpb]
    \begin{tabular}{lcccc}
    %\hline
    \textbf{Method}          & \textbf{Citeseer} & \textbf{Cora} & \textbf{Pubmed}  & \textbf{Wikipedia}\\
    \hline
    Uniform                  &  7.7  &  13.0  &  18.0   & 28.7  \\ \hdashline
    GDR (Uniform)                &    50.6 (+42.9)   &   71.8 (+58.8)  &  73.2 (+55.2)   &  31.4 (+2.7)  \\ \hline
    Projection                &  61.8  &  59.0  &   72.0  & 32.5  \\
    RF         &  60.3  &  58.9  &  68.8  &  {\bf50.8} \\
    SVM           &  61.1   & 58.0   &  49.9   & 31.0  \\
    MLP          & 57.0 & 56.0 &  70.7 &  43.0 \\ 
    \hdashline
    GDR (Projection)                 & 70.4 (+8.7) & 79.7 (+20.7) & 75.8 (+3.8) & 36.9  (+4.4)\\ 
    GDR (RF)          & 70.5 (+10.2) & 78.7 (+19.8) & 72.2 (+3.2) &  50.8  (+0.0) \\
    GDR (SVM )             & 70.3 (+9.2)  & 81.2 (+23.2) & 52.4 (+2.5)  & 41.9  (+10.8)\\ 
    GDR (MLP)                    & 69.7(+12.7) & 78.5 (+22.5) & 75.5 (+4.8)  & 40.5 (-2.5) \\
 \hline
    Planetoid          & 64.7    & 75.7 &   72.2  & - \\
    GCN         & 70.3     & 81.1 &  79.0 & 39.2  \\
    \hdashline
GDR (GCN)                    & {\bf 70.8} (+0.5) & {\bf 82.2} (+1.1) & {\bf 79.4} (+0.4)  & 39.5 (+0.3) \\
    \hline
    \bigskip
    \end{tabular}
    \caption{
    Percentage classification accuracy before and after application of relabelling by GDR for various classifiers.  We present the improvement of GDR on the uniform prediction (which ignores features). We also consider four supervised classifiers (which learn from features without the graph): projection, RF, SVM and MLP.  For RF, we use a maximum depth of $20$; for SVM, we set $C=50$; for MLP, we implement the same architecture as GCN ($d_1=16$-unit hidden layer, $0.5$ dropout, $200$ epochs, $0.01$ learning rate, $L^2$ loss function). Finally, we compare with two semi-supervised graph classifiers: GCN~\cite{Kipf2016} and Planetoid~\cite{Yang2016}. The numbers in brackets record the change in accuracy accomplished by applying GDR on the corresponding prior classifier.}
        \label{tbl:results}
\end{table}

\subsubsection*{Classification performance}

Table~\ref{tbl:results} summarises the percentage classification accuracy before and after the application of GDR for various prior classifiers.  
Our main observation is that for the Citeseer, Cora and Pubmed datasets,
\textit{a posteriori} relabelling by GDR improves significantly the classification accuracy of all prior classifiers, achieving comparable accuracy to GCN without the need for semi-supervised learning through the graph, i.e., GDR(RF) improves GCN by 0.2\% on Citeseer; GDR(SVM) improves by 0.1\% on Cora; GDR(Projection) falls short by 3.2\% on Pubmed.  Note that GDR consistently outperforms Planetoid, another top-ranking semi-supervised graph classification method, on these datasets. 

Our results also provide insight into the relative importance and alignment~\cite{qian2019quantifying} of the features and the graph for the purpose of classification. 
The comparison of the Uniform assignment (which has no information from the features) with the relabeling induced by GDR(Uniform) reveals large improvements in performance for all three datasets, from 43\% in the case of Citeseer to 59\% in the case of Cora. This observation underlines the fact that the graph contains useful information for classification even in the absence of feature information.  
On the other hand, adopting information from the features alone (without the graph) through supervised classifiers also induces large increases in performance in these three datasets (from 46\% in Cora to 54\% in Citeseer and Pubmed). 
When applying GDR to these feature-based classifiers we observe that the maximum synergistic improvement is obtained for Cora, whereas the improvement of GDR is smaller for Pubmed, signalling a lower alignment of the graph with the features and the ground truth, as discussed previously in~\cite{qian2019quantifying}.

The Wikipedia dataset constructed in~\cite{qian2019quantifying} has low alignment between features, graph and ground truth. The heuristic behind this difference is simple: the hyperlinks in Wikipedia articles (citations) are not necessarily aligned with the content of the categories of the articles, i.e., an article from a mathematician will be linked to its country of birth. Hence, in this case, the graph contains information which is incongruous with the features and ground truth, so that the GCN performs worse than using features-only classifiers, such as MLP and especially RF. 
Interestingly, the random forest classifier for the Wikipedia dataset is the only case where GDR is not able to reclassify any nodes, suggesting that no information from the graph is able to improve the RF classification purely based on features. Similarly, the application of GDR to the output of GCN only induces marginal improvement, underscoring the fact that GCN has already made use of the information in the graph.
In general, however, the application of the re-classification step (GDR) always increases the original accuracy (expect for MLP in Wikipedia), yet by different amounts, depending on the structure of the prior.

Our results above thus show that GDR achieves state-of-the-art classification accuracy, just by diffusing the class label information explicitly on the graph without the need for graph-based inference. 

\section{Diffusive GCN (diff-GCN)} \label{sec:MT_vs_GCN}

 A natural extension to the GCN architecture is to include an explicit diffusion within the learning phase of the GCN architecture with the aim to increase the classification accuracy. A straightforward approach is to replace the one-step transition matrix $\widehat A$ in each layer of~\eqref{eq:fullgcn} by the transition matrix of the diffusion, $e^{-tL}$.
 The two-layer GCN model in~\eqref{eq:fullgcn} then becomes
\begin{align}
\begin{bmatrix} * %\widehat{H}
\\  H_\text{diff-GCN} \end{bmatrix} = \text{softmax}\left(e^{-tL} \
    \text{ReLU} \left(e^{-tL} 
    \mathcal{X}\,
    W^0 \right) W^1 \right)\, ,
    \label{eq:diff_gcn}
\end{align}
where the matrices $W^0(\widetilde{X})$ and $W^1(\widetilde{X})$, and the diffusion time parameter $t(\widetilde{X})$ are inferred during the learning phase from the training set. Since $t$ is the same for all nodes in the graph, it can be thought of as a global scale for the convolution, which allows for additional flexibility in using the scales in the graph (beyond the one-step transitions in standard GCN).
We call this construction \textit{diffusive GCN} (diff-GCN). 

The results of diff-GCN show a slight improvement on all the benchmark datasets, as shown in  Table~\ref{tbl:gcnimprovements_exp}. The largest improvement in the accuracy was seen for the Wikipedia dataset, where the features were not aligned with the graph topology~\cite{qian2019quantifying}. This suggests that using an adaptable, optimised continuous-time diffusion transition operator offers higher flexibility when exploring more subtle relationships between features and graph. 

\begin{table}[htpb]
    \begin{tabular}{lcccc}
    \textbf{Model}          & \textbf{Citeseer} & \textbf{Cora} & \textbf{Pubmed}  & \textbf{Wikipedia}\\    \hline
    GCN              &  70.3   & 81.1 & 79.0 &  34.1\\ 
    diff-GCN             & 71.9 & 82.3 & 79.3 & 45.9 \\ \hline
    \\
    \end{tabular}
    \caption{ 
    Percentage classification accuracy of GCN and its extension diff-GCN, which has an explicit diffusion operator~\eqref{eq:diff_gcn}. 
    }
    \label{tbl:gcnimprovements_exp}
\end{table}

\section{Extensions to directed graphs} \label{sec:directed_graph}

In many cases of interest, the pairwise relationships between samples are \textit{asymmetric}, e.g., following \textit{vs.}\ being followed on a social network~\cite{beguerisse2012twitter}, or the highly directed synaptic connectivity between neurons~\cite{bacik2016flow}. Such asymmetry can be highly informative of the structure of the dataset. Clearly, in such cases, the ensuing graph is directed, with a non-symmetric adjacency matrix, $A \neq A^T$. We now present extensions of the GDR and diff-GCN frameworks to carry out classification tasks with directed graphs exploiting the natural connection of our methods with diffusive dynamics.

Extending GDR and diff-GCN to directed graphs follows closely from above, yet one needs to consider specifically the diffusive process on such graphs. Note that unless the directed graph is strongly connected, the stationary state of the diffusive dynamics is concentrated on the `dangling nodes' (i.e., the nodes without outgoing edges). 
To avoid such trivial asymptotic behaviours, it is customary to consider an associated diffusive process that contains a `reinjection' (also known as Google teleportation) that guarantees ergodicity~\cite{lambiotte2014random}. 
The transition matrix of this associated process is:
\begin{align}
    P_\text{dir}(A,\alpha) = \alpha \ D_\text{out}^+ A + 
    \Big( (1- \alpha) I_N + 
\alpha \ \text{diag}\left(\mathbbm{1}_0(A \boldsymbol{1}) \right) \Big) \frac{\boldsymbol{1} \boldsymbol{1}^T}{N}\, . 
\end{align}
where $D_\text{out}^+$ denotes the pseudo-inverse of the out-degree matrix, 
$D_\text{out}=\text{diag}(A \boldsymbol{1})$, and 
$\mathbbm{1}_0(A \boldsymbol{1})$ 
is the indicator vector for the dangling nodes (i.e., nodes with vanishing out-degree).
This process evolves on the directed edges of the graph with probability~$\alpha$ (customarily chosen to be $\alpha=0.85$), and transitions uniformly to any node in the graph with probability $(1-\alpha)$~\cite{page1999pagerank,Chung2005,beguerisse2012twitter}. The probability at the dangling nodes is reinjected with probability 1.
Clearly, the transition matrix is row-stochastic (i.e., $P_\text{dir} \boldsymbol{1} =\boldsymbol{1}$). The Perron left eigenvector $\boldsymbol{\varphi}$ of this matrix (i.e., $\boldsymbol{\varphi}^T P_\text{dir}= \boldsymbol{\varphi}^T$) is the well-known Pagerank~\cite{page1999pagerank}. See~\cite{Chung2005} for more details. 

To remain consistent with the undirected cases presented above, we use here the symmetric part of the combinatorial directed Laplacian~\cite{Chung2005}
\begin{align}
\label{eq:Ldir}
    L_\text{dir}(A, \alpha) = \Phi - \frac{\Phi P_\text{dir} + {P_\text{dir}}^T \Phi}{2}\, ,
\end{align}
where $\Phi=\text{diag}(\boldsymbol{\varphi})$.
Clearly, if  $A \neq A^T$ then $L_\text{dir}(A, \alpha) \neq L_\text{dir}(A^T, \alpha)$. (If $A=A^T$ and $\alpha=1$, we recover $L_\text{dir}(A, 1) = L_\text{dir}(A^T, 1) = L$ for the undirected case.)
With these definitions, the extension of our methods to directed graphs is straightforward. 
Given a directed graph of binary relationships between the samples with $N \times N$ adjacency matrix $A \neq A^T$, we have the following:

\subsubsection*{GDR on directed graphs}
The GDR algorithm remains as described in Section~\ref{sec:GDR-section}, except that we use $L_\text{dir}$ instead of $L$ to compute the overshootings in Eq.~\eqref{eq:overshoot}.

\subsubsection*{GCN and diff-GCN on directed graphs}

The original formulation of GCN did not pursue the use of directed graphs~\cite{Kipf2016}. Here we have implemented a directed version of GCN, which applies the same equations~\eqref{eq:fullgcn}--\eqref{eq:gcn_graph_loops} to an asymmetric adjacency matrix $A$.

Similarly, the only change for the application of diff-GCN to a directed graph is to substitute $L$ for $L_\text{dir}$ in Eq.~\eqref{eq:diff_gcn}.

\subsubsection*{Augmented (bi-directional) GCN and diff-GCN on directed graphs}

In a directed graph, the adjacency matrix $A$ is associated with a transition matrix for forward propagation, whereas its transpose $A^T$ is associated with backward propagation of the process. In some cases, important features about the dataset can be extracted from \textit{both} forward and backward information propagation~\cite{cooper2009,beguerisse_vangelov_RMST2013}.

The GCN architecture can naturally accommodate learning through convolutions operating in parallel on different graphs~\cite{Kipf2016}.  
We have applied this principle to create an augmented GCN model where the forward and backward propagation are unfolded to enable the inference of separate parameters for each channel $A$ and $A^T$ in parallel.  This strategy allows the flexibility to learn the most relevant convolution operators for the directed graph and its reverse, as if they were two different graphs defined on the same set of nodes.

To write the augmented GCN model for a directed graph with adjacency matrix $A \neq A^T$, we first define the forward and backward symmetrised transition matrices, which have the same form as~\eqref{eq:gcn_graph_loops}:
\begin{align}
\label{eq:forward_backward}
    \widehat{A}_\text{fw} & = \bar{D}_\text{out}^{-1/2} \bar{A} \; \bar{D}_\text{out}^{-1/2} \\
       \widehat{A}_\text{bw} & = \bar{D}_\text{in}^{-1/2} \bar{A}^T \bar{D}_\text{in}^{-1/2}\, ,
\end{align}
where $\bar{A}=A+I_N$, $\bar{D}_\text{out}=\text{diag}(\bar{A} \boldsymbol{1})$, and
$\bar{D}_\text{in}=\text{diag}(\bar{A}^T \boldsymbol{1})$. We compile these two matrices in the augmented matrix $\widehat{\mathcal{A}}_{N \times 2N}=\begin{bmatrix} \widehat{A}_\text{fw} \, \widehat{A}_\text{bw} \end{bmatrix}$. 

The augmented two-layer model can be compactly represented in matrix form using the Kronecker product:
\begin{align}
\begin{bmatrix} * %\widehat{H}
\\  H_\text{aug-GCN} \end{bmatrix} = 
\text{softmax}\left( 
\widehat{\mathcal{A}} \, \left (\text{ReLU} 
\left(\widehat{\mathcal{A}} \, \left(\mathcal{X} \otimes I_2 \right ) 
\mathcal{W}^0
\right) \otimes I_2
\right ) \mathcal{W}^1
\right)\, ,
    \label{eq:augmented_gcn}
\end{align}
where $I_2$ is the identity matrix of dimension 2 and the parameters to be inferred for each of the models (forward and backward) are compiled in the augmented matrices
$\mathcal{W}^0_{2F \times d_1} = \begin{bmatrix} W_\text{fw}^0 \\ W_\text{bw}^0 \end{bmatrix}$ and
$\mathcal{W}^1_{2d_1 \times c} = \begin{bmatrix} W_\text{fw}^1 \\ W_\text{bw}^1 \end{bmatrix}$. 

Similarly, the augmented diff-GCN model has the same form as~\eqref{eq:augmented_gcn} with the substitution of the operator $\widehat{\mathcal{A}}$ for an augmented matrix containing the explicit forward and backward diffusion operators based on $A$ and $A^T$: 
\begin{align}
\widehat{\mathcal{A}}_{N \times 2N} \mapsto
    \mathcal{E}_{N \times 2N}=\begin{bmatrix} 
    e^{-t \, L_\text{dir}\left(A\right)} \enskip 
    e^{-t \, L_\text{dir}\left( A^T \right)} 
    \end{bmatrix},
\end{align}
where $L_\text{dir}(A)$ is defined in~\eqref{eq:Ldir} and we fix $\alpha=0.85$.

\subsection{Application to the directed Cora dataset}

Citations are not reciprocal. By their own nature, citation networks like the ones considered in this paper are therefore highly directional. 
It is worth noting that the publicly available citation graphs for the Cora, Citeseer and Pubmed datasets are all \textit{undirected}. In our computations above (Tables~\ref{tbl:results}~and~\ref{tbl:gcnimprovements_exp}), we have used the available undirected graphs to facilitate comparisons of our results with published work.
However, the question remains as to the impact of the directionality of citations in the classification task. 

To examine this point, we have returned to the original Cora dataset and construct the directed graph $A_\text{dir}$ of citations from the raw data\footnote{The directed graph from Cora is available on~\url{https://github.com/barahona-research-group/GDR}.}. 
We then use this directed graph (and its transpose) within the GDR and diff-GCN frameworks. 

\begin{table}[htpb]
    \begin{tabular}{lccc}
    & \textbf{Undirected} & \textbf{Directed (fw)} & \textbf{Directed (bw)} \\ 
     \textbf{Method} & $A$  & $ A_\text{dir}$ & $ A_\text{dir}^T$    \\ 
    \hline
    GDR (Projection)     & 79.7   & 62.1 & 64.6 \\
    GDR (RF)   & 78.7 & 58.0  & 57.6   \\ 
    GDR (SVM)   &  81.2 & 63.6  & 62.1 \\ 
    GDR (MLP)   &  78.5 & 57.3  & 56.4 \\ 
    \hline  \\
    \end{tabular}
    \caption{
    Accuracy of GDR using the undirected, directed, and reverse directed graphs of the Cora dataset.}
    \label{tbl:directed}
\end{table}

The results of GDR for the Cora dataset for the directed $A_\text{dir}$, its transpose $A_\text{dir}$ and the undirected version $A$ are compared in  Table~\ref{tbl:directed}. In all cases, we see a reduction of accuracy for the directed graphs, as compared to the previous undirected algorithm. This indicates that the classes benefit from considering both directions (who cites and who is cited) in the inference of scientific topic.  

The results of GCN and diff-GCN using the undirected, directed, reverse directed, and bi-directional (i.e., augmented) versions of the graph classification algorithms are presented in Table~\ref{tbl:bi-directed}.
Again, we observe that the use of both directions for label diffusion improves the classification. The best performance ($83\%$ accuracy) is achieved with the augmented diff-GCN, due to the flexible use of both directions of the diffusion together with the optimised scale provided by the diffusion time $t$. 

\begin{table}[htpb]
    \begin{tabular}{lcccc}
    & \textbf{Undirected} & \textbf{Directed (fw)} & \textbf{Directed (bw)} & \textbf{Augmented (fw,bw)} \\ 
    \textbf{Method} & $A$  & $ A_\text{dir}$ & $ A_\text{dir}^T$ & $\begin{bmatrix}  A_\text{dir} \,  A_\text{dir}^T \end{bmatrix}$    \\ 
    \hline
        GCN    & 81.1 & 67.4 & 79.8 &  79.9 \\
    diff-GCN  & 82.3 & 80.3 & 81.7 & \bf{83.0}\\ 
\hline \\
    \end{tabular}
    \caption{Accuracy of GCN and diff-GCN using the undirected, directed, reverse directed, and bidirectional (augmented) graphs of the Cora dataset. }
    \label{tbl:bi-directed}
\end{table}

\section{Discussion}\label{sec:conclusion}

In this paper, we have introduced two methods for semi-supervised classification that make use of a graph capturing the relationships between samples. First, we presented GDR, a re-classification algorithm that leverages the diffusion on the graph to relabel any prior  probabilistic classification on the samples. Our numerical experiments on three benchmark datasets with several prior classifiers show that GDR consistently improves the accuracy of the original classifier 
without the need for a high quality prior classification. Hence GDR can be used as a post-processing tool to refine class assignments by taking into account a graph that formalises additional information about relationships between samples. 

In addition, our results show that GDR provides comparable results to GCN, the state-of-the-art semi-supervised graph classification method, yet without the need for graph-biased inference. 
Deep learning methods (such as GCN) infer the classifier from the features and the graph simultaneously.  However, this assumes that the feature or label mixing on the graph should be homogeneous. Instead, our GDR relabeling allows us to take a node-centric view of classification: the amount of information gathered from the graph to reclassify a node is different for each node. 
GDR side-steps the use of graph-based deep learning architectures by carrying out the classification in two steps: a feature-based classification followed by a reclassification incorporating the graph information. This approach also allows to establish the relative importance (and the alignment) of the information contained in the features and the graph (as shown by the Wikipedia example in Table~\ref{tbl:results})~\cite{qian2019quantifying}.

As a second method based on diffusion, we introduced diff-GCN, a simple extension of the GCN algorithm that embeds the explicit diffusion operator with time as a hyper-parameter to be inferred. Our results show that this additional flexibility improves slightly the original GCN method on the benchmark cases. The dynamics-based viewpoint also allows us to introduce extensions of diff-GCN for directed graphs based on ergodicised diffusions on such graphs. We showed that allowing both the forward and backward diffusions to take place on the graph improved the accuracy to  $83\%$ in the directed graph for one of our datasets (Cora). This is one of the highest accuracies on the Cora dataset, just falling short of the $83.5\%$ recently set by Dual GCN~\cite{Zhuang2018}, which uses an additional long range convolution.

The examples in this paper show that using graph diffusion in conjunction with classification algorithms can provide natural extensions and interpretations to deep learning architectures. Although we have concentrated here on classification problems, similar ideas could be used for dimensionality reduction and multiscale unsupervised clustering, where graph-based methods also provide interesting links with spectral methods in clustering and machine learning~\cite{lambiotte2014random,MSC,zijing2017,zijing2019}. These directions will be the object of further study.

 \subsection*{Code and data availability}
The Python code to compute GDR is available at \url{https://github.com/barahona-research-group/GDR}.

\section*{Acknowledgments} 
We thank Dominik Klein, Hossein Abbas, Paul Expert, Yifan Qian, Asher Mullokandov and Sophia Yaliraki for valuable discussions. We acknowledge EPSRC funding through award EP/N014529/1 via the EPSRC Centre for Mathematics of Precision Healthcare.

\bibliographystyle{abbrv} 
\bibliography{refs}      

%\medskip
% The data information below will be filled by AIMS editorial staff
%Received xxxx 20xx; revised xxxx 20xx.
%\medskip

\end{document}